# Concept and the implementation of a tool to convert industry 4.0 environments modeled as FSM to an OpenAI Gym wrapper

Kallil M. C. Zielinski
kallil@alunos.utfpr.edu.br
UTFPR - Federal University of Technology Paraná
Pato Branco, PR, Brazil

Marcelo Teixeira
marceloteixeira@utfpr.edu.br
UTFPR - Federal University of Technology Paraná
Pato Branco, PR, Brazil

Richardson Ribeiro
richardsonr@utfpr.edu.br
UTFPR - Federal University of Technology Paraná
Pato Branco, PR, Brazil

Dalcimar Casanova
dalcimar@utfpr.edu.br
UTFPR - Federal University of Technology Paraná
Pato Branco, PR, Brazil

## ABSTRACT

Industry 4.0 systems have a high demand for optimization in their tasks, whether to minimize cost, maximize production, or even synchronize their actuators to finish or speed up the manufacture of a product. Those challenges make industrial environments a suitable scenario to apply all modern reinforcement learning (RL) concepts. The main difficulty, however, is the lack of that industrial environments. In this way, this work presents the concept and the implementation of a tool that allows us to convert any dynamic system modeled as an FSM to the open-source Gym wrapper. After that, it is possible to employ any RL methods to optimize any desired task. In the first tests of the proposed tool, we show traditional Q-learning and Deep Q-learning methods running over two simple environments.

## 1 INTRODUCTION

In the era of the fourth industrial revolution, frequently noted as Industry 4.0, one key for Cyber-Physical Production Systems is the ability to react adaptively to dynamic circumstances of production processes [11][20][29]. This ability leads to the necessity of developing acting components (actuators), like robots, for example, with some capacity of self adaptation and learning, such that they can modify their behavior according to the experience acquired through interactions with each other and with the environment. Actuators with such characteristics are called intelligent agents [37].

This rises a question on how industry 4.0 components can be programmed in such a way that they recognize variations in the environment, and autonomously adapt themselves to act accordingly, in a concurrent, safe, flexible, customized, and maximally permissive way. In conjunction, those features make hard the task of programming such industrial controllers agents, as usual paradigms for software development become inappropriate.

From the automation perspective, the behavior of a system can be described by its evolution over time. When this evolution comes from signals observed in physical equipment and devices, it usually has an asynchronous nature over time, which ends defining how this system can be represented by the model. The class of systems that share this asynchronous characteristic is called *Discrete Event Systems* (DESs) [5], whose modeling is in general based on *Finite State Machines* (FSMs).

However, FSMs face significant limitations when modeling large and complex dynamics with SEDs. Advanced features, such as context recognition and switching, and multiphysics phenomena, are difficult to be expressed by ordinary FSMs. They are usually associated with large and intricate models that not rarely have to be built by hands, which challenges both modeling and processing steps. With the increasing demand for flexibility, the ordinary FSM-based modeling methods have become insufficient to express emerging phenomena of industrial processes, such as dynamic context handling [27].

This paper is based on the idea that an actuator agent, can sense the environment and be controlled safely, as usual, by using the conventional control synthesis methods. However, they can additionally complement their actions with an operating agent that observes the plant under control and gathers to it experiences about its interaction with the environment. In other words, the observer acts as a recommender for the events that are eligible to occur. Based on the experiences calculated by the agent, it seeks to adapt its behavior and control the components in such a way that desired tasks are performed as intended, but in an optimized, adaptive, way.

This concept of adaptability in conjunction with large complex industrial environments, whose predictability is practically non-existent and the main objective is the optimization of results, makes industrial environments ideal for applying model-free methods and all modern RL concepts. These environments have a large number of states, can be viewed as an Markov process and are oriented towards a final objective.

However, the first difficulty in direct application of reinforcement learning (RL) methods over the industry 4.0 is the lack of that environments in order simulate the industry process. Although the industrial process can be modeled as a DES, this model is not suitable for direct application of RL methods. To RL methods we need a simulated environment where the agent can interact and learn, explore new states and transitions in order to reach a good solution for the desired objective.

The existence of several environments is a crucial point to success in the application of RL methods, especially the deep based ones, over different knowledge areas such as games [19], classic robotics [13], natural language processing [17], computer vision [3], among others. So, our main objective here is to provide the same diversity of environments to industry 4.0, initially focused on the coordination of industry components, but not limited to that.

In this way, this paper presents the concept and the implementation of a tool that allows to convert any dynamic system modeled as a FSM to the open-source *Gym* environment [4]. The result is a complex computational structure, obtained through simple, condensed and modular design, that is apt to receive modern RL treatment. Once modeled as a Gym, them DES structure can be viewed as a Markov Decicion Process (MDP) and processed by means of RL [32], *Q-learning* [35], or any other *Deep* RL approach [19] without lots of problem-specific engineering. The result is a complex computational structure, obtained through simple, condensed and modular design, that is apt to receive modern RL treatment.

Gym is a toolkit for developing and comparing reinforcement learning algorithms. It makes no assumptions about the structure of your agent, and is compatible with any numerical computation library, such as *Pytorch* [22] or *Tensorflow* [1]. Once modeled as a gym, the DES structure can be viewed as a *Markov Decision Process* (MDP) and processed by means of RL [32], *Qlearning* [35], or any other *Deep* RL approach [19] without lots of problem-specific engineering.

Structurally, the manuscript is organized as follows: a brief literature overview is presented in Section 2; Section 3 discusses the background; Section 4 compares details of RL and DES; Section 5 introduces the main results; a case study is presented in Section 6; and some conclusions and perspectives are presented in Section 8.

## 2 STATE OF THE ART

Reinforcement learning applications in industry are nothing new. However in [8] the author argues that implementing such methods in real industry environments often is a frustrating and tedious process and academic research groups have only limited access to real industrial data and applications. Despite the difficulties, several works have been carried out. The work of [15] reviews applications both in robotics and in industry 4.0.

From a perspective of industrial environments the [8] it is the work that presents an idea closer to the one proposed here in this article. The authors designed a benchmark for the RL community to attempt to bridge the gap between academic research and real industrial problems. Its open source based on OpenAI Gym is available at https://github.com/siemens/industrialbenchmark. At Gym official page [21] there is a few Gym third party environments that resembles an industrial plant or robotics.

However, the general use and application of RL in industry are still limited by the static, one-at-a-time, way in which the environment is modeled, with rewards and states being manually reconfigured for each variation of the physical process. This approach can be replaced, to some extent, by a DES-based strategy in order to map those variations more efficiently.

In fact, while part of the behavior of industrial systems is classically continuous in time, the events happen in a discrete setting and variations are essentially stochastic. Therefore, the ability for mapping industrial processes as DESs can allow handling events and variations more efficiently via RL. DES-based approaches has been classically used in industry for both modeling and control [25]. However, cyber-physical features of processes are still difficult to be captured by ordinary theories and tools for DES [27].

In this paper, we claim that any industrial environment that can be modeled as a DES, can be automatically transformed in a OpenAI Gym wrapper which bridges the gap between real world applications and RL area. We remark that, in most cases, components of a DES can be described by simple, compact, and modular DES models that can be combined automatically to represent the entire system. Therefore, the task of obtaining the system model is never actually a burden to be carried entirely by the designer, which can be decisive in RL applications.

## 3 BACKGROUND

### 3.1 Emerging industrial systems

Intensive data processing, customized production, flexible control, autonomous decisions. Much has been discussed about these assertions in recent years, in the context of *Industry 4.0* (I4.0).

I4.0 emerges from evolution of centralized production systems, based on embedded microprocessors, to the distributed *Cyber-Physical Systems* (CPSs) [6, 7, 16]. A CPS is responsible for the fusion of real and virtual worlds, therefore indispensable link with modern systems. Technically, it integrates components or equipment with embedded software, which are connected to form a single networked system. This integration model leads to the acquisition and traffic of a large volume of data, processed locally and made available to other components via the Internet [2].

A CPS links production processes to technologies such as *Big Data*, *Internet of Things*, *Web Services*, and *Computational Intelligence*, which together allow the setup of highly flexible processes, promisingly more efficient, cost-effective, and on demand for each user profile [10]. In practice, the I4.0 principles disrupt the usual production model consolidated so far, based on centralization and large factories, creating decentralized, autonomous, interconnected, and intelligent production chains that promise to support the industry of the future.

It s conceivable, however, to imagine how many possible obstacles separate the I4.0 from its consolidation as a *de facto* industrial revolution. Safety, interoperability, performance, risks, etc., impose severe restrictions to its practice. Also the technical infrastructure of a I4.0-based system is not necessarily compatible with the current methods of control and automation, requiring an additional level of integration.

The conversion method proposed in this paper can be seen as a tool to improve interoperability among I4.0, process, and expert methods. We focus on the event level of the process, which is discussed in the following.

### 3.2 The discrete nature of industrial systems

The modeling of a system is justified by several reasons, including the fact that it is not always possible, or safe, act experimentally over its real structure. Moreover, a model allows engineers to abstract irrelevant parts of the system, facilitating its understanding.

If a system has a discrete nature and its transitions are driven by sporadic events, it is called a Discrete Event System (DES).

DES is a dynamic system that evolves according to physical signals, named *events*, that occur in irregular and unknown time intervals [5]. This contracts with dynamic systems that evolve continuously in time.

### 3.3 Formal background on SEDs

Differently from continuous-time systems, that can be naturally modeled by differential equations, DESs are more naturally represented by Finite State Machines (FSMs). An FSM can be formally introduced as a tuple $G = \langle \Sigma, Q, q^\circ, Q^\omega, \rightarrow \rangle$ where: $\Sigma$ is finite set of events called *alphabet*; $Q$ is a finite set of *states*; $Q^\omega \subseteq Q$ is a subset of marked states (in general associated with the idea of complete tasks); $q^\circ \in Q$ is the *initial state*; and $\rightarrow \subseteq Q \times \Sigma \times Q$ is the *transition relation*.

Sometimes, it is convenient to expose $G$ in a usual graphical convention, although this view may not be illustrative for large models. Figure 1 shows the graphical view of an FSM modeling a simple machine with only two states.

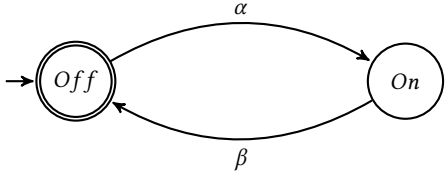

**Figure 1: Graphical layout of an FSM**

In this case, a transition between any two states $q, q' \in Q$, with the event $\sigma \in \Sigma$, is denoted by $q \xrightarrow{\sigma} q'$. In this example, $Q = \{on, off\}$, $\Sigma = \{\alpha, \beta\}$, $q^\circ = off$, and $Q^\omega = \{off\}$ means that tasks are completed only when the machine is turned off, which in this case coincides with the initial state. From now forward, FSMs are exposed graphically or, when it is too large to layout, we mention only its number of states and transitions, which are acceptable measures for its dimensioning.

When a DES is formed by a set $J = \{1, \cdots, m\}$ of components, which is quite often, each component can be modeled by a different FSM $G_j$, $j \in J$ and combined afterward by *synchronous composition*. This allows the entire system to be designer modularly, which can be decisive for large-scale systems. The result is a global, combined, behavior (also called the *plant*), where all components work simultaneously without any external restriction. For this reason, the plant composition is also known as *open-loop* plant.

Consider two FSMs, $G_1 = \langle \Sigma_1, Q_1, q_1^\circ, Q_1^\omega, \rightarrow_1 \rangle$ and $G_2 = \langle \Sigma_2, Q_2, q_2^\circ, Q_2^\omega, \rightarrow_2 \rangle$. The synchronous composition of $G_1$ and $G_2$ is defined as $G_1 \| G_2 = \langle \Sigma_1 \cup \Sigma_2, Q_1 \times Q_2, (q_1^\circ, q_2^\circ), Q_1^\omega \times Q_2^\omega, \rightarrow \rangle$, where elements in the set $\rightarrow$ satisfy the following conditions:

- $(q_1, q_2) \xrightarrow{\sigma} (q_1', q_2')$ if $\sigma \in \Sigma_1 \cap \Sigma_2, q_1 \xrightarrow{\sigma} q_1'$, and $q_2 \xrightarrow{\sigma} q_2'$;
- $(q_1, q_2) \xrightarrow{\sigma} (q_1', q_2)$ if $\sigma \in \Sigma_1 \setminus \Sigma_2$ and $q_1 \xrightarrow{\sigma} q_1'$;
- $(q_1, q_2) \xrightarrow{\sigma} (q_1, q_2')$ if $\sigma \in \Sigma_2 \setminus \Sigma_1$ and $q_2 \xrightarrow{\sigma} q_2'$.

The synchronous compositions merges (synchronizes) events shared between $G_1$ and $G_2$, and it interleaves the others. A transition that does not follow any of these rules is said to be undefined, which in practice means it is disabled.

### 3.4 Modeling the system

The first step to map a process as a DES is to identify which components (or subsystems) are expected to be modeled. Subsystems are then be individually modeled by an FSM and composed afterward. At this step, the designer observes only the constraint-free behavior of components, disregarding details about their posterior coordination.

Considering a DES formed by a set $J = \{1, \cdots, m\}$ of components, so that each component is modeled by an FSM denoted $G_j$, $j \in J$. Then

$$G = \|_{j \in J} G_j$$

is said to be the *plant* model.

*3.4.1 Example of plant modeling.* In order to illustrate the modular way a DES can be modeled, consider a simple example of two transmitters, $T_1$ and $T_2$, sharing a communication channel $C$, as in Fig. 2(a). $T_1$ and $T_2$ can be respectively modeled by the FSMs $G_{T_1}$ and $G_{T_2}$ in Fig. 2.

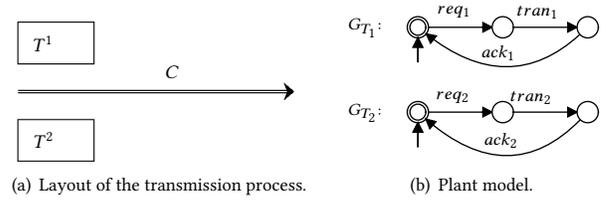

(a) Layout of the transmission process.   (b) Plant model.

(c) Composed plant model $G = G_{T_1} \| G_{T_2}$.

**Figure 2: Example of a concurrent transmission system.**

When seen as a DES, the following events are observable throughout the process:

- $req_1$ and $req_2$: are request messages arriving for transmission in $T^1$ and $T^2$, respectively;
- $tran_1$ and $tran_2$: model the start of transmission in $T^1$ and $T^2$, respectively; and
- $ack_1$ and $ack_2$: reset the channel $C$ for new transmissions.

From the observation of this event set, one can construct a plant model that represent the dynamic behavior of the channel based on its evolution over the state-space. A proposal for modeling each transmitter is shown in Fig 2(b), where each model is composed of three states, meaning respectively the process idle, waiting transmission, and transmitting. The plant $G = G_{T_1} \| G_{T_2}$ has 9 states and 18 transitions, and it is displayed in Fig 2(c).

Remark that, even for this simple example of two transmitters, the plant model $G$ requires a substantially large state-space to expose and unfold all possible sequences for the system. Yet, the designer has never actually faced this complexity, as the most complex FSM is modeled with only 3 states and model $G$ emerges from an automatic composition.

As the behavior of each transmitter (and consequently of $G$) is unrestricted, i.e., it does consider channel limitations neither the sharing of $C$ with other transmitters, they have to be restricted to some extent. This is approached next.

### 3.5 Restricting the system

When components of a DES are modelled by FSMs, their composition leads to a plant $G$ that expresses the unrestricted system behaviour. In practice, the plant components need to follow a certain level of coordination for them to operate concurrently and behave as expected.

For this purpose, an additional structure called a *restriction*, here denoted by $R$, has to be composed with the plant. A restriction can be seen as a prohibitive action which is expected to be observed in the system behavior. In summary, as the plant has been modeled by a composition of constraint-free subsystems, we now disable some of its eligible events, adjusting them to cope with the restrictions.

In practice, a restriction

$$R = \|_{i \in I} R_i,$$

for $i \in I = \{1, \cdots, n\}$, can also be expressed by automata and composed automatically to the plant $G$. This leads to the so-called *closed-loop* behavior, in this paper denoted by $K = G\|R$.

*3.5.1 Example of modeling restrictions.* For the transmission system example in Fig 2, consider that $C$ has capacity for transmitting only one message at a time. In this case, the behaviors of the transmitters have to be restricted with respect to the channel capacity. An FSM that models such a limitations presented in Fig. 3.

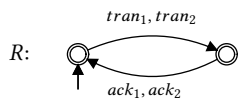

**Figure 3: Mutual exclusion restriction $R$ for the channel $C$.**

That is, $R$ imposes mutual exclusion to transmissions in the channel. It allows both transmitters to start transmission (events $tran_1$ and $tran_2$ from the initial state), but, as soon as one of them occupies the channel, the other is prohibited to transmit until an acknowledge ($ack_1$ or $ack_2$) is received (both $tran_1$ and $tran_2$ are disabled in the non-initial state).

### 3.6 Closed-loop modeling

Remark, therefore, that controlling a DES plant $G$ relies, first of all, on obtaining a model $R$ that reflects correctly the expected requirements. This is a design tasks that in this paper is assumed to be well-defined.

From the composition $K = G\|R$, one obtains an FSM $K$ that models the closed-loop system behavior, i.e., the system behavior under the control of $R$. For the previous example of the transmission system in Fig 2, for instance, $K = G\|R$ has 8 states and 14 transitions, and it is displayed in Fig 4.

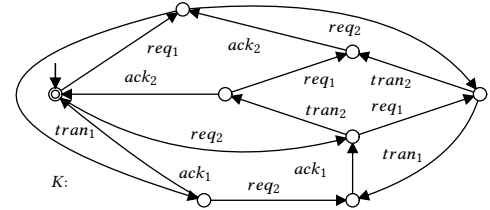

**Figure 4: Graphical view of $K = G\|R$.**

Model $K$ can be converted into implementable hardware language, for practical use [23].

### 3.7 Controllability of events

Remark that $K$ can be seen as an preliminary version of a control logic for the plant $G$. However, from an industrial point of view, $K$ is expected to have some additional properties before it can be implemented. It is expected, for example, that it differentiates controllable and non-controllable events.

By now, we have assumed that all events in $G$ can be handled under control. In practice, however, some events may occur in an involuntary fashion, so that they cannot not be directly handled. These events are called *uncontrollable* and not considering them may violate the control consistency, situation when the controller commands a certain action that cannot be reproduced physically.

A communication breakdown or a signal dropout, for example, are samples of uncontrollable events. In Fig 2, $i_{Req}$ and $i_{Ack}$, $i = 1, 2$ are also uncontrollable, as one cannot decide whether or not a message arrives or a transmission is confirmed. They are observable and expected to occur, eventually, but they cannot be handled in advance by the controller, and therefore must be kept free to occur.

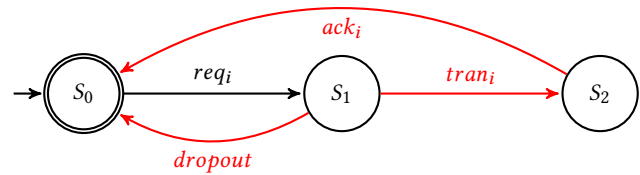

**Figure 5: Addiction of a dropout signal in the plant models**

Formally, the idea of controllability of events can be presented by partitioning the set of events of an FSM, such that $\Sigma = \Sigma_c \cup \Sigma_u$ turns to be the alphabet resulting from the union of *controllable* ($\Sigma_c$) and *uncontrollable* ($\Sigma_u$) events. Then, mathematical operations can be defined to extract from a model $K$ its sub-model that respects the impossibility of disabling events in $\Sigma_u$. This is the kernel of, for example, some control synthesis methods such as the Supervisory Control Theory [24] and its several extensions.

In this paper, controllability will be considered and it plays an essential role in the results to be derived later in Section 5 involving RL. However, we do not go through the synthesis processing step, leaving this to the engineer's discretion. In fact, our goal here is to associate a controllability-aware model $K$ with an adaptive-aware RL method, so that the controllability itself can be exploited either before or after the proposed conversion of $K$ to gym.

If on one hand, processing synthesis over $K$ (before the RL treatment) assigns robustness to the control system, on the other hand it reduces its chances for flexibility. Our approach works for both strategies, but we opted by abstracting the synthesis step in order to illustrate the potential of our approach for flexible, customizable, control.

## 3.8 Reinforcement Learning

Reinforcement learning is a computational paradigm in which an agent wants to increase its performance based in the reinforcements that receive on interacting with an environment, learning a policy or a sequence of actions [9][31]. To do that, the agent that acts in an environment perceives a discrete set of states $S$, and realizes a set $A$ of actions. In each time step $t$, the agent can detect its actual state $s$ and, according to this state, choose an action to be executed, which will take it to another state $s'$. To each state-action pair (s,a) there is a reinforcement signal given by the environment, $R(s, a) \longrightarrow \mathbb{R}$ to the agent when executing and action $a$ in state $s$.

The most traditional way to formalize the reinforcement learning consists on using the concept of *Markov Decision Process* (MDP). An MDP is formally defined by a quadruple $M = \langle S, A, T, R \rangle$, where:

- $S$ is a finite set of states in the environment;
- $A$ is a finite set of actions that the agent can realize;
- $T : S \times A \longrightarrow \prod(S)$ is a state transition function, where $\prod(S)$ is a probability distribution over the set of states $S$ and $T(s_{t+1}, s_t | a_t)$ defines the probability of realizing the transition from state $s$ to state $s + 1$ when executing an action $a_t$, and
- $R : S \times A \longrightarrow \mathbb{R}$ is a reward function, which specifies the agent's task, defining the reward received by an agent for selecting action $a$ being in a state $s$.

The successful application of modern RL was mainly demonstrated in board games (e.g. backgammon [34], go [28] and Atari games [19]). In all these cases, the environments are MDPs or partially observable MDPs (POMDPs). Figure 6 shows a simple example of an MDP with 5 states and 6 actions, in which state 1 is the initial state and 5 is the terminal (or objective) state.

In the MDP above, we can assume that the quadruple $M = \langle S, A, T, R \rangle$ is consisted of these elements:

- $S$ ={1,2,3,4,5};
- $A$ ={Read a book, Do a project, Publish a paper, Get a Raise, Play Video Game, Quit};

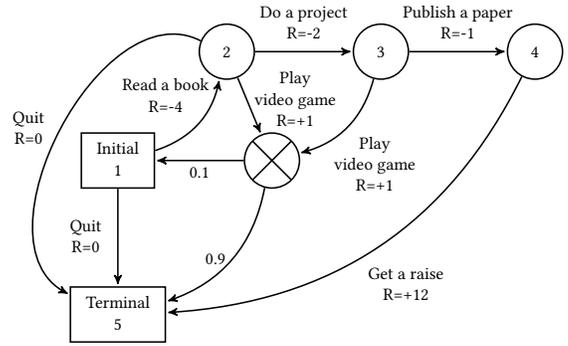

**Figure 6: Example of an MDP of a real life academic.**

- The state transition function consists of all probabilities being 100% except for states 2 and 3 for action Play Video Game, which have a probability of 10% of returning to state 1 and 90% of going into terminal state 5. This transition is denoted by ⊗ symbol.
- The reward table is shown in Table 1, which consists on all of the rewards received by the agent on doing the action related to the table's columns being on the state related to the table's line. Note that all cells with a NaN (not a number) indicates that the transition is not possible on the MDP.

In order to an agent maximize his rewards, he need to learn that the cumulative reward over time can only be maximized when temporary punishments, that is, negative rewards, are accepted. In the example above, for the agent to get a raise, first he needs to read a book, do a project and publish a paper, and all of these previous actions give the agent a negative reward, but in the end, his reward has added a positive 12, so all of the hard work was worth it. Therefore, the agent need to take in account not only immediate rewards, but also possible future rewards. A single episode $e_{MDP}$ can be described as a sequence of states, actions and rewards:

$$e_{MDP} = s_0, a_0, r_0, s_1, a_1, r_1, \cdots, s_{n-1}, a_{n-1}, r_{n-1}, s_n \quad (1)$$

where $s_i$ represent the i-th state, $a_i$ the i-th action and $r_i$ the i-th reward. The total future reward at any time point $t$ is given by:

$$R_t = r_t + \gamma r_{t+1} + \gamma^2 r_{t+2} + \cdots + \gamma^{n-t} r_n \quad (2)$$

where $\gamma \in (0, 1)$ represents the discount factor and models how strongly the agent takes future rewards into account. Values close to 0 will represent a short-sighted strategy as higher-order terms for rewards in the distant future become negligible. If the environment is deterministic, $\gamma$ can be set to 1 as the same actions always result in the same rewards [14].

With that, it is possible to define the aptitude of an agent that learns with reinforcement as the aptitude to learn a policy $\pi^*$ : $S \times A$ which maps the actual state $s_t$ in a desired action, being able to maximize the accumulated reward over time, describing the agent's behavior ([9]).

A good strategy to try maximize the future reward can be learned through the state-action value function, or also the Q function. It specifies how good it is for an agent to perform a particular action in a state with a policy $\pi$. We can define the Q values as follows:

$$Q^\pi(s,a) = \mathbb{E}_{s'}\left[r + \gamma \mathbb{E}_{a'}\ \pi(s')[Q^\pi(s',a')]\right] \quad (3)$$

where $Q^\pi(s,a)$ is the Q value (quality value) of a policy $\pi$ of doing an action $a$ in a state $s$, and $\mathbb{E}$ is the expected value. This equation is also called the Bellman Equation.

|  | **Read a Book** | **Do a Project** | **Publish a Paper** | **Get a Raise** | **Play video game** | **Quit** |
|---|---|---|---|---|---|---|
| **State 1** | -4 | NaN | NaN | NaN | NaN | 0 |
| **State 2** | NaN | -2 | NaN | NaN | +1 | 0 |
| **State 3** | NaN | NaN | -1 | NaN | +1 | NaN |
| **State 4** | NaN | NaN | NaN | +12 | NaN | NaN |
| **State 5** | NaN | NaN | NaN | NaN | NaN | NaN |

**Table 1: Reward table for Figure 6 MDP.**

There is a wide variety of algorithms of RL, such as Q-Learning [36], H-Learning [33], Dyna [30], Sarsa [31] Deep Q [19], among others.

In this paper we make use of two well known algorithms in RL: Q-learning and Deep Q network.

### 3.9 Q-learning

The Q-learning algorithm [35] has attracted lots of attention for its simplicity and effectiveness. This algorithm allows to establish a policy of actions in an autonomous and iterative way. It can be shown that the Q-Learning algorithm converges to an optimal control procedure, when the learning hypothesis of state-action pairs Q is represented by a complete table holding the information of the value from each pair. The convergence occurs in both deterministic and non-deterministic Markov Decision Process.

The basic idea of Q-learning is that the learning algorithm learns an evaluation function under all the state-action pairs $S \times A$. The Q function provides a mapping in the form $Q : S \times A \longrightarrow V$ where $V$ is the value of the expected utility of executing an action $a$ in a state $s$. Since the agent's partitions of both state space and action space do not omit relevant information, once the optimal function is learned, the agent will know which action will result in a better future reward in all of the states.

Considering the Bellman Equation (Equation 3), if the policy $\pi$ tends to be an optimal policy, the term $\pi(s)$ can be considered as $argmax_a Q^\pi(s,a)$. Then we can rewrite the Equation 3 as:

$$Q^\pi(s,a) = \mathbb{E}_{s'}\left[r + \gamma max_{a'}\ \pi(s')[Q^\pi(s',a')]\right] \quad (4)$$

From the last equation, in a discrete state space the Q-learning uses an online off-policy update, so the equation of the Q-values can be formulated as:

$$Q^\pi(s,a) = Q^\pi(s,a) + \alpha\left(r + \gamma max_{a'} Q(s',a') - Q(s,a)\right) \quad (5)$$

where $Q(s,a)$ is the Q-value of executing an action $a$ in a state $s$, $\alpha$ is the learning rate of the training, and $maxQ(s',a')$ is the highest value in the Q-table in the line related to the state $s'$ next to the state $s$, that is, the action of the next state that has better returns according to the Q-table. This Q function represents the discounted expected reward under taking an action $a$ when visiting state $s$, and following an optimal policy since then. The procedural form of the algorithm is shown below.

---

**Algorithm 1:** Q-learning procedure

1. Receives $S$, $A$, $Q$ as input;
2. For each $s, a$, set $Q(s,a) = 0$;
3. **while** *Stop Condition == False* **do**
4.     Select action $a$ under policy $\pi$;
5.     Executes $a$;
6.     Receive immediate reward $r(s,a)$;
7.     Observe new state $s'$;
8.     Update $Q(s,a)$ according to Equation 5;
9.     $s \leftarrow s'$;
10. **end**
11. Return $Q$;

---

Note that the stop condition can be executing a specific number $n$ of steps in a single episode, reach a terminal state, among other methods. And once all state-action pairs are visited a finite number of times, it is guaranteed that the method will generate estimates $Q$, which converge to a value $Q^*$ [36]. In practice, the action policy converge to the optimal policy in a finite time, although slowly. Moreover, it is possible to learn the ideal control directly, without modelling the transition probabilities or the expected rewards present on the Equation 4, this makes potential the Q-learning use in Discrete Event Systems.

However, Equation 5 always chooses the action with the highest Q-value for that specific state $s_i$. Let us say that, at first, the agent is in the MDP initial state $s_0$, and take an action $a_1$ which gives him a good reward, so upon updating the Q-table, there is a value higher than 0 in $Q(s_0, a_1)$. The agent does not know that if he takes an action $a_0$ in the initial state $s_0$ gives him a better reward than taking action $a_1$, but he will always chooses action $a_1$ in that state because it is the maximum value on the Q-table, and therefore will not have knowledge that there are better actions to be taken. To improve that, so that the agent explores all state-action pairs in table Q, the use of epsilon-greedy as a policy $\pi$ of Algorithm 1 is required.

The epsilon-greedy policy detailed in Algorithm 2, shows that the objective is to switch between exploration (choose random action) and exploitation (choose best action) according to a value $\epsilon$ between 0 and 1 that defines the probability of choosing a random

Concept and the implementation of a tool to convert industry 4.0 environments modeled as FSM to an OpenAI Gym wrapper action to explore all the actions in the environment. Upon receiving the current state $s_i$, the Q-table $Q$ and $\epsilon$ as the input, the algorithm generates a random number between 0 and 1. If this number is higher than $\epsilon$, then the policy will choose the action with the highest Q-value in that state $s_i$ according to $Q$. But if the number is lower or equal than $\epsilon$, then it will choose a random action to take in the environment.

---

**Algorithm 2:** Epsilon greedy policy

1 Receives $s_i$, $Q$,$\epsilon$ as input;
2 Generate random number $x$ between 0 and 1;
3 **if** $x > \epsilon$ **then**
4     Choose action with highest Q-value in $Q(s_i)$;
5 **else**
6     Choose random action;
7 **end**

---

In the end, Q-learning is a tabular method which is very useful in many systems, however, its effectiveness is limited to environments with a reduced number of states and actions, making an exhaustive search for all possible state-action pairs contained in the environment. If we consider an environment with an extremely high number of states, and in each state, there is a high number of possible actions to be taken. It would be a waste of time and memory to explore all of the possible state-action pairs. A better approach would be to use an aproximation function with $\theta$ parameters in which $Q(s, a; \theta) \approx Q^*(s, a)$.

To do that, we can use a neural network with parameters $\theta$ to approximate the Q-values for all possible state-action pairs [14]. This approach was created in [19] with the objective of making an AI to learn how to play atari games, and it was called Deep Q Network (DQN).

### 3.10 Deep Q Network

The goal of the deep Q network is to make unnecessary the exhaustive search for all state-action pairs of the environment. Because in some cases, there can be a huge number of states. An example is a servo motor's posicion, which can assume values between 0 and 180 degrees, however, this is not a discrete value, but a continuous variable that can assume many values in this interval.

Another case is an atari game, that has in image of 210 x 160 pixels, and a RGB color system in which each channel can be varied from 0 to 255. If we consider that each state is a single possible frame, the number of possible states in the environment is approximately $210x160x255x255 \approx 5.57e^{11}$. In the Atari case, a single pixel does not make much difference, so both images are treated as a single state, but is still necessary to distinguish some states. In [18], the Deepmind researchers have tested a convolutional neural network with seven different atari games, and in [19] they made some optimizations in the same neural network and in the environments which became possible the training of 49 different games. In both cases, the authors used a convolutional neural network to extract the game state, and then use dense layers to get an approximation function of the Equation 5.

In DQN, the loss function at iteration $i$ that needs to be optimized is the following:

$$L_i(\theta_i) = \mathbb{E}_{s,a,r,s'}\left[\left(\hat{y}_i - Q(s,a;\theta_i)\right)^2\right], \quad (6)$$

where $\hat{y}_i = r + \gamma max_{a'}Q(s', a'; \theta')$, and $\theta'$ denotes the parameters of the target network. This target network is the same as the original network, but as the original network is updated every step, the target network is updated every N steps, and each of these updates corresponds to the target network copying the parameters of the original network.

These 2 separate networks are created because in some environments, like the Atari games, there is a lot of consecutive states very similar to each other, since these 2 steps can be a single pixel changing, so there is a lot of correlation between them. In this case, the values of $Q(s, a; \theta)$ and $Q(s', a'; \theta)$ are very similar, which means that the neural network cannot distinguish well between both states [14], causing the training to be really unstable.

Another important ingredient to reducing the correlation among steps is the *Experience Replay*, where the agent accumulates a buffer $\mathcal{D} = t_1, t_2, ..., t_i$ with experiences $t_i = (s_i, a_i, r_i, s_{i+1})$ from many episodes. And the network would be trained by sampling from $\mathcal{D}$ uniformly at random instead of directly using the current samples. And the loss function can be expressed as:

$$L_i(\theta_i) = \mathbb{E}_{s,a,r,s' \sim u(\mathcal{D})}\left[\left(\hat{y}_i - Q(s,a;\theta_i)\right)^2\right] \quad (7)$$

Figure 7 [18] shows the neural network structure used in [19] since the extraction of the game state until the Q value of each action to the related state. In this case, the actions to be used are related to each button (or combination of buttons) of an Atari controller.

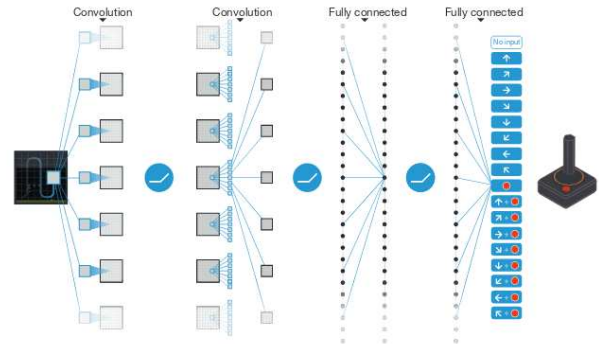

**Figure 7: Neural network used for training the games.**

## 4 RELATIONSHIP BETWEEN MDP AND FSM AND THE IMPLICATION OVER RL BEHAVIOR

By comparing the structure of an MDP and an FSM, the following can be highlighted to help sustaining the propositions in Section 5:

- A state of an MDP is equivalent to a state of an FSM, therefore, also the initial state of an MDP is equivalent to the

initial state of an FSM, and a terminal (marked) state of an MDP coincides with a terminal (objective) state of an FSM.
- An action in an MDP coincides with an event that triggers a transition in an FSM;
- In an MDP, an action is executed by an agent, while an event is expected to be handled by a controller in an FSM modeling a DES. Both events and actions have controllability issues, i.e., some events are expected to be uncontrollable, and some actions cannot be entirely decided by the agent. Take the MDP in Figure 6 as an example. In states 2 and 3 the agent can make the action "Play video game", but the subsequent state is undetermined, since there is a probability of 90% to reach the terminal state 5, and 10% to reach the initial state 1. To adapt this as a DES, we may call the action "Play video game" as a controllable event, that evolves the DES to the intermediary state in the MDP, the one with the two transition probabilities. And these two transitions (0.9 and 0.1) would be labeled with uncontrollable events that evolve the DES to the initial state represented by the MDP state 1, or to the terminal state 5. In summary, all actions in an MDP are events in a DES, but only actions corresponding to controllable events can be taken by the agent, the others are unknown. In this work, we consider to split the set of actions $A$ into the sets $A_c$ and $A_u$, defining respectively controllable and uncontrollable actions, such that $A_c \subseteq A$, $A_u \subseteq A$, and $A_u \cup A_c = A$ follow by construction.

To make the understanding clearer, Table 2 shows a summary of the relationship between a MDP and a DES.

Despite apparent similarities, a DES model does not include natively a reward processing system to evaluate the agent's actions. This prevents it to be directly exploited for control optimization purposes, which is an important feature in I4.0 environments. In contrast, RL is a reward-aware by construction, which makes it closer to the I4.0 needs, but it lacks immediate resources for safety-aware DES modeling and control.

In this paper, we claim that both approaches are useful, can be combined to some extent, but they are not because of the lack of integration tools. Based on this claim, we consider to address basic control requirements (those related to safety) directly on the FSM level (see section 3.6), and optimization requirements by converting the semi-controlled DES into a RL model. The RL reward processing step is described in the following.

### 4.1 Reward processing analysis

In RL, two notions of rewards can be considered: immediate and delayed. In I4.0 context, a delayed reward would be, for example, the conclusion of a manufacturing step, leading to a profit. In terms of modeling, this profit should to be set manually for all events (or actions) that lead to a terminal state.

Many real-life applications of RL use delayed rewards, so that RL naturally solves the difficult problem of correlating immediate actions with the delayed returns they produce. Like humans, RL algorithms sometimes have to wait a while to collect return from their decisions. They operate in a delayed return environment, where it can be difficult to understand which action leads to a specific outcome over many time steps.

On the other hand, immediate rewards are also possible to implement and they can be positive or negative. A positive immediate reward, in I4.0, can be for example part of a final product being manufactured (e.g. the lid of a kettle). Although the part is not related entirely to possible profit, when finished it can bring indirect benefits to the production plant (e.g. it can release specific machines to steps that lead to the full product manufacturing).

Differently, a negative immediate reward is any action (i.e., an event in a DES) that consumes resources (e.g. energy, time, raw material) and does not result in a complete product, or part of it. The more common case is the association of small negative values to events that cause waste of time or energy (e.g. a robot move). In other cases, such as a broken machine (i.e. an uncontrollable event) can have very high negative reward, but the probability for this event to occur is very low. In this case, the RL method needs to learn through simulation the chances for this event to occur, and if this results in positive rewards by the end of a manufacturing process (delayed reward). In the broken machine example, the immediate negative reward can be associated with the cost of repair, while the delayed reward can be associated with the profit of producing a number $n$ of assets.

It is common, for a I4.0 process modeled as DES, to have positive rewards only on the events that reach final states, while the intermediate events are all negative and can be associated with time costs, raw material, power cost, labor, etc.

## 5 DES MODEL CONVERSION TO A GYM WRAPPER

In this section, we detail the proposed conversion of DES models into trainable gym wrapper. The resulting procedures and codes can be accessed at [12].

For a better understanding of the methods and functionalities of a gym wrapper, we first detail the main characteristics of this type of environment. Then, we describe features of the FSM that serves as input to the learning environment. Finally, we present a detailed methodology and conversion steps, which are exposed and released in the form of a computational tool.

### 5.1 Gym wrappers

Gym is a toolkit for developing and comparing reinforcement learning algorithms. It makes no assumptions about the structure of the agent, and is compatible with any numerical computation library, such as TensorFlow or Pytorch [4].

The goal of a Gym wrapper is to interact with a modeled environment by exploring its state-space. By observing states, the agent learns how good it would be to perform specific actions in each state. This allows it to choose actions that maximize rewards by the end of a task, thus obtaining an optimal action policy.

The *step* method in our gym class returns four values that represent all the necessary information needed for training. They are detailed in the following:

- **Observation (Object)**: a specific object of the environment that represents the observation of the current state in that environment;
- **Reward (Float)**: amount of reward received by the previously chosen action;

Concept and the implementation of a tool to convert industry 4.0 environments modeled as FSM to an OpenAI Gym wrapper

| Discrete Event System (DES) | Markov Decision Process (MDP) | Symbol as DES | Symbol as MDP | Elements Relationship |
|---|---|---|---|---|
| Set of states | Is the set of states of an MDP | $Q$ | $S$ | $Q = S$ |
| Initial State | Is the initial state of an MDP | $q^\circ$ | $s_0$ | $q^\circ = s_0$ |
| Set of marked states | Is the set of terminal states of an MDP | $Q^\omega$ | $S_m$ | $Q^\omega = S_m$ |
| Controllable Events | Is all of the actions that the agent has control over it, that is, it can decide whether or not to take that action. | $\Sigma_c$ | $A_c$ | $\Sigma_c = A_c \subseteq A$ |
| Uncontrollable Events | In MDP, the uncontrollable events represents the set of uncontrollable actions, in which the RL agent does not have control, e.g. the transition probabilities in Figure 6. The transition probabilities in a DES are uncontrollable events. In this paper, we consider that the user specifies probabilities for all uncontrollable events and, if not specified, there is equal probabilities for them to trigger. For example, in a state $x$, there is 2 enabled uncontrollable events, so there is 50% of chances for each to trigger. | $\Sigma_u$ | $A_u$ | $\Sigma_u = A_u \subseteq A$ |
| Transitions | A transition in a DES is also a transition in an MDP, which evolves the environments from a state to another. Although in an MDP there is some transitions followed by probabilities of going to different states, in this paper we consider that all of the transitions of the generated MDP have 100% of probability to go to the next state, i.e., once a transition is activated, we will know what will be the next state in the environment. | $f$ | $T$ | $f = T$ |

Table 2: Relationship between DES and RL environments

- **Done (Boolean)**: identifies if the task for the environment is complete. This can happen when the MDP is in a terminal state or simply performing a specific $N$ number of steps. After, this variable becomes *True*, a method called *reset* is invoked to return the environment to its initial state and reset all rewards to 0;
- **Info (Dictionary)**: diagnoses information used for debugging. It can also be eventually useful for learning.

From the observation object, we can choose actions to be taken by the agent, that go through all states in the environment, capturing the required information. Note that all actions are assumed to be available for the agent, in all possible states of the environment. In most gym environments, available at the original gym website, there is also a *render* function that allows the user to see the visual representation of the environment. This can be a screenshot on an atari game, a drawing of a chess configuration board, etc.

In this paper, as we working specifically with FSMs, all environments are represented as automata. The *render* function will return the automaton of the environment, with initial, current, and last state specified in the observations.

### 5.2 DES system to a gym wrapper

The first step to convert a DES model into a gym wrapper is designing the plant $G$ and specifications $R$ in a suitable modeling software. Here we use *Supremica* [39], a design-friendly tool that includes resources for both modeling and control tasks, besides to allows composing, simulating, and checking the correctness of FSM models.

Upon composition, one obtains the FSM $K = G \| R$, i.e., the behavior expected for the system under control, exactly as projected by the engineer. The model $K$ could be further exploited for control purposes, as for example in terms of controllability of its events, nonblockingness, etc., which is quite straightforward in control engineering practice. However, for the purposes in this paper, we consider to keep $K$ as it is, i.e., including immediate control actions projected via engineering, for it to be further refined using RL. This differs, to certain extent, to the control practice, which not usually associate the robustness of control with the smoothness and sensitivity of RL techniques. We claim this as a novelty of our proposal.

With the pre-controlled FSM $K$ on hands, useful information can be extracted and converted to an MDP. For example, the set of states, initial state, set of marked states, events (controllable and uncontrollable), and transitions (see Table 2). In practice, these information were exported as a XML structure, and parsed to a gym class in order to create an MDP environment.

Then, this MDP environment is a class containing all information specified by the DES, following the relationship in Table 2. The attributes of this new class are: states, events, and transitions of the input DES. There is also some attributes that specify the initial state, set of terminal states, and controllability of events. In a standard gym wrapper, it is considered that all actions are possible in all states. In this work, however, we modify this assumption to consider only actions possible in a given state. This coincides with FSM models that have their state-transition formalized as a partial function. For a given state $s_i$, we may call $A^i$ the subset of possible actions in that state, whith $A^i \subseteq A$.

In addition to the information described in the Table 2, a RL environment also requires the reward structure for taking a specific action in a given state. In this work, we consider that all actions taken by the agent return an immediate reward representing a profit (e.g. the production of a workpiece) or a loss (e.g. spent time, consumed energy, raw material) in our system. Thus, for the environment to work, it is necessary a set of rewards $R$, in which an element $r_i \in R$ is specified for each action in the environment. $R$ is provided by as a parameter to the *reset* method in the form of a list. As default, all actions receive a loss of $-1$, informing that every action has a loss. The loss type does not need to be specified in the system, as it is a generalization of any desired optimization objective. All default rewards can be converted into a positive reward (a profit) or negative (loss). These rewards depend on the system to be modeled, and there are no predefined rules. However, in most cases, positive rewards are defined for actions related to production, as it represents profit in industrial environments.

On the other hand, there are some features of DESs that are not directly converted in MDP, such as the controllability of events. Although both types of events make an FSM evolve from a state to another, we have to consider that the uncontrollable events cannot be activated or disabled by control. These events are triggered by a probability function (see the example of "Play video game" action in Figure 6). Therefore, we consider that the set of uncontrollable events ($\Sigma_u$) of a DES turns into a set of uncontrollable actions $A_u$ in an MDP. These actions cannot be chosen by a RL agent and their activation will be determined by an associated uniform probability. The set of probabilities, $P$, in which each $p_i \in P$ is the probability of the system to trigger an action $a_i$, can also be passed by parameter to the *reset* function.

By default, if $P$ is not specified, then the policy $\pi$, in the exploration mode, chooses randomly between all the possible actions, both controllable and uncontrollable, to be triggered in a given state. That is, if in a given state $s_i$ we have two possible uncontrollable actions ($a_1$ and $a_2$) and one controllable action ($a_3$), then there is 33% of chances for $a_1$ and $a_2$ to be chosen, and 33% for $a_3$. In Figure 10, the action $ack_1$ is an example of an uncontrollable action that always triggers. This default configuration reflects the situation where the DES designer is unaware of the activation frequency for uncontrollable transitions.

If $P$ is specified only for some uncontrollable actions, but not for all, then we verify if these actions trigger. If not, we do epsilon-greedy with the remaining actions in order to trigger one of them. Take as an example the model in Figure 5, where we can specify a probability of 1% to the dropout action. In that case, this action is evaluated first. After that, other possible actions are evaluated by the exploration or exploitation strategies of the policy $\pi$.

REMARK 1. *In case of choosing exploitation, we need to pick up in that state the action with the highest Q-value, as long as the action is controllable, as the agent cannot choose uncontrollable actions.*

REMARK 2. *Triggering or not uncontrollable actions, based on probabilities or the exploration, is a matter only during the training phase. Upon training, the agent chooses only controllable actions with the highest Q-value, while uncontrollable ones are triggered physically.*

REMARK 3. *Controllable events are not triggered by probabilities, since the control agent decides whether or not to take it. In this way, the RL agent learns an optimal policy on training, and it chooses the action to take in a given state based on the best possible reward. But the selected action cannot be uncontrollable, because the agent can not choose to either activate it or not.*

From these remarks, there are some considerations to introduce. For example, the exploration-exploitation policies need to be extended. Epsilon-greedy for example, turns into a new algorithm called *Controllable Epsilon-greedy policy*, presented in Algorithm 3. In the new implementation, the policy function receives the possible transitions $A^i$ enabled from a state, their controllable nature, and the list of transition probabilities $P$.

First, the algorithm stores all possible uncontrollable actions of $A^i$ in a list of possible uncontrollable transitions called $A_u^i$, i.e., $A_u^i = A^i \cap A_u$. Similar idea applies for controllable transitions, i.e., $A_c^i = A^i \cap A_c$. If $A_u^i \neq \emptyset$, then we iterate over $A_u^i$, in which $A_u^i[j] = a_{uj}^i$ and verify if $P[a_{uj}^i] > 0$. If so, then we store the tuple

---

**Algorithm 3:** Controllable Epsilon greedy policy

1 Receives $s_i, Q, \epsilon, A^i, P, A_u$ as input;
2 $A_u^i = A^i \cap A_u$;
3 $A_c^i = A^i \cap A_c$;
4 **if** $A_u^i \neq \emptyset$ **then**
5     Create list $\Delta$;
6     **while** *Iterate over $A_u^i$ such that $A_u^i[j] = a_{uj}^i$* **do**
7         **if** $P(a_{uj}^i)$ **then**
8             Generate random number $\zeta_i$ between 0 and 1;
9             $\Delta \leftarrow (a_{uj}^i, P(a_{uj}^i), \zeta_i)$;
10            Remove $a_{uj}^i$ from $A^i$;
11         **end**
12     **end**
13     **if** $\Delta \neq \emptyset$ **then**
14         Shuffle $\Delta$;
15         **while** *Iterate over $\Delta$* **do**
16             **if** $\zeta_i > P(a_{uj}^i)$ **then**
17                 Choose action $a_{uj}^i$;
18             **end**
19         **end**
20     **end**
21 **end**
22 **if** $A_c^i \neq \emptyset$ **then**
23     Generate random number $x$ between 0 and 1;
24     **if** $x > \epsilon$ **then**
25         Choose $A_c^i$ action with highest Q-value in $Q(s_i)$;
26     **else**
27         Choose random action in $A^i$;
28     **end**
29 **else**
30     Choose random action in $A^i$;
31 **end**

---

$(a_{uj}^i, P(a_{uj}^i), \zeta_i)$ in a list called $\Delta$, where: $a_{uj}^i$ is the uncontrollable possible action of index $i$ in $A_u^i$; $P(a_{uj}^i)$ is the probability of occurrence of action $a_{uj}^i$; and $\zeta_i$ is a randomly generated number for each action $a_{uj}^i$.

If the list $\Delta$ is not empty, i.e., there is a probability higher than 0 for events in the set $A_u^i$, then we shuffle the list $\Delta$ and iterate over it. If, for any of the elements, there is a $\zeta_i$ higher than $P(a_{uj}^i)$, we choose the action $a_{uj}^i$. The shuffle in list $\Delta$ is necessary to simulate situations where a number $n > 1$ of uncontrollable events is possible to occur, and it is obviously unknown which one occurs first (e.g. broke machine or power shutdown).

In case none of the uncontrollable events is triggered, the events in $a_{uj}^i$ are removed from the set $A^i$. If $A_c^i$ is not empty, a random number between 0 and 1 is generated. If this number is less than $\epsilon$, the agent chooses the action with the highest Q-value in $A_c^i$, otherwise it chooses a random possible action.

The diagram in Figure 8 summarizes the steps for making the gym wrapper adaptations from a DESs, starting from the initial modelling of the system until the effective application of RL algorithms.

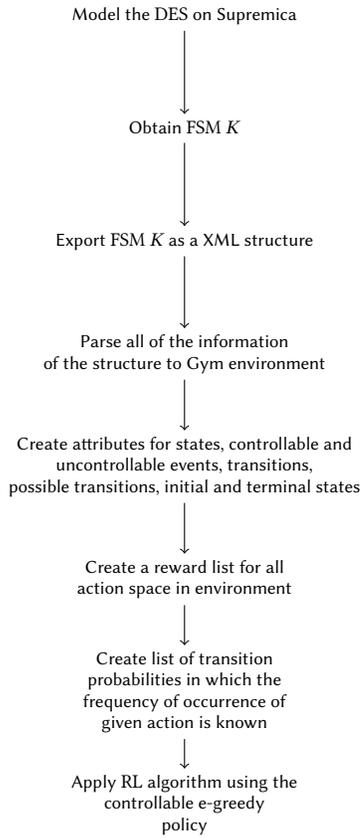

Figure 8: Conversion methodology.

Table 3 summarizes the extra sets that, together with the DES model parameters in Table 2, are used in the gym wrapper.

| Symbol | Definition |
|---|---|
| $R$ | List of rewards for the DES's events. |
| $P$ | List of probability for the DES's events. This is optional for the Gym environment |
| $A^i$ | Set of possible actions in state $s_i$. |
| $A^i_c$ | Set of possible controllable actions in state $s_i$. |
| $A^i_u$ | Set of possible uncontrolable actions in state $s_i$. |

Table 3: List of extra sets for the MDP conversion

## 6 EXAMPLES OF USE OF THE ENVIRONMENT

Two examples are presented in this section to illustrate the applicability of the developed tool. The first is a DES with two machines and an intermediary buffer to stock workpieces. The second example is the system with the 2 concurrent transmitters anticipated in Section 3.4. Both examples are separated in 3 steps:

(i) Modelling the system, to obtain the FSM $K$;
(ii) Conversion of the DES model to an MDP environment;
(iii) Apply the reinforcement learning algorithm, discuss, and review the results.

Also, we will consider that, according to [26], we may define that both of our environments are discrete, completely observable, and static. Both examples and the conversion tool are available at GitHub repository [38].

### 6.1 Two machines with intermediate buffering

As a study case, consider the manufacturing system shown in Figure 9, composed by 2 machines, $M_1$ and $M_2$, and an intermediary buffer $B$. Machine $M_1$ picks up a workpiece (event $a_1$), manufactures, and delivers it in the buffer $B$ (event $b_1$). Machine $M_2$ picks up the workpiece from $B$ (event $a_2$), manufactures, and removes it from the system (event $b_2$). The buffer is considered to support only one workpiece at a time. It is also assumed the possibility for the machines to break.

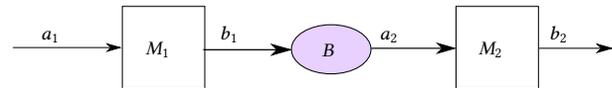

Figure 9: Example of manufacturing system.

*6.1.1 Modelling of the system:* In this example, we consider the events $a_i$ as controllable, since we can avoid a machine to start, and the events $b_i$ as uncontrollable, since we cannot force a machine to conclude a job. We also consider that events $c_i$ model break events for the machines $M_i$ and they are uncontrollable, while events $r_i$ model the respective repair and they are controllable. The plant models for $M_1$ and $M_2$ are presented in Figure 10.

Events $a_i$ indicating that machine $M_i$ is operating, while $b_i$ leads the automaton to its initial state, also marking a completed task. While operating, the machine can crash, which is triggered by the event $c_i$, and repaired with an event $r_i$, which leads the machine back to the initial state. We design uncontrollable transitions in red, in the hope that this could facilitate identification, while the others are kept in black.

Assume that the control objective is to avoid underflow and overflow in the buffer, by disabling some events occur. For this purpose, we compose to the plant model the restriction modeled by the FSM $E_1$, shown in Figure 11.

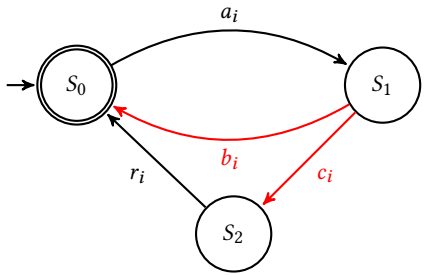

**Figure 10: Plant models $G_i$ for machines $M_i$, $i = 1, 2$.**

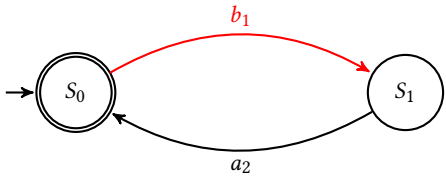

**Figure 11: Underflow and overflow restriction $R$.**

Model $R$ controls both underflow and overflow of workpieces in the buffer, by respectively disabling $a_2$ in the initial state (when the buffer is empty in state $S_0$), and disabling $b_1$ otherwise (when the buffer is already full in state $S_1$).

By composing plant and restriction models, one obtains the FSM $K = G_1 \| G_2 \| R$, which models the expected behavior for the system under control. $K$ has 18 states and 42 transitions, and it is shown in Figure 12.

*6.1.2 Converting DES to MDP.* After parsing the model through the XML structure, in order to reveal the structure of $K$ as seen by the RL environment, we call the function *render* implemented in the gym environment and the result is presented in Figure 12, which shows the initial state of the environment, right after calling the *reset* function (note the green painted state).

Also, Figure 13 shows the environment after making an action $a_1$ (note the purple painted transition), which makes the automaton evolve to state $S_3$. In state 0, none of the machines are operating. After an action $a_1$, the FSM moves to state 3 where machine $M_1$ is operating until an uncontrollable action $b_1$ or $c_1$ triggers, meaning that machine $M_1$ either delivers a workpiece to the buffer or crashes, respectively. The remaining evolution follows similar reasoning for every triggered action.

Before a RL algorithm can be effectively applied, it remains to be constructed a reward list for all actions in the environment, and a probability list of all uncontrollable actions. We may consider that all movements that lead a machine to NOT produce workpieces return a loss of -1 as the default value; and all movements leading to a workpiece production returns a profit of 10, in this case, only action $b_2$.

We also consider that a machine crash produces a loss of -4, because of the wasting time and costs to repair. We consider that a machine crashing has 5% chance of happening. Thus, the reward list is shown in Table 4, where the values marked with * are default.

| Action set $A$ | Reward set $R$ | Probability set $P$ (%) |
|---|---|---|
| $a_1$ | -1* | not applicable |
| $b1$ | -1* | not specified |
| $a2$ | -1* | not applicable |
| $b2$ | 10 | not specified |
| $c1$ | -4 | 5 |
| $c2$ | -4 | 5 |
| $r1$ | -1* | not applicable |
| $r2$ | -1* | not applicable |

**Table 4: Rewards for the manufacturing system example.**

Is it worth remembering that the controllable actions are chosen by the control agent to trigger, so there is no need to adopt probabilities for their occurrences and we show this as *not applicable* in the table. There is also some uncontrollable actions that are not related with any standard of frequency of occurrence. For them, the probability is set as *not specified*. We are now in position to apply the RL algorithm.

*6.1.3 Applying RL and reviewing the results.* In this example we use the Q-learning algorithm to represent the training of the environment. We also considered that the episode ends after the agent performs 60 actions. For the action selection, we use the policy implemented in Algorithm 3, and train the environment for 100 episodes. The resulting Q-table is shown in Table 5.

|  | $a_1$ | $a_2$ | $b_1$ | $b_2$ | $c_1$ | $c_2$ | $r_1$ | $r_2$ |
|---|---|---|---|---|---|---|---|---|
| **St. 0** | 13.14 | - | - | - | - | - | - | - |
| **St. 1** | 20.69 | - | - | 21.80 | - | 0.51 | - | - |
| **St. 2** | 5.99 | - | - | - | - | - | - | 7.24 |
| **St. 3** | - | - | 15.74 | - | 3.90 | - | - | - |
| **St. 4** | - | - | 23.03 | 24.15 | 8.99 | 5.94 | - | - |
| **St. 5** | - | - | 8.29 | - | -1.76 | - | - | 12.20 |
| **St. 6** | - | - | - | - | - | - | 10.58 | - |
| **St. 7** | - | - | - | 17.83 | - | -0.76 | 15.89 | - |
| **St. 8** | - | - | - | - | - | - | 0.36 | 1.25 |
| **St. 9** | 17.63 | 18.61 | - | - | - | - | - | - |
| **St. 10** | 24.72 | - | - | 26.71 | - | 5.01 | - | - |
| **St. 11** | 8.75 | - | - | - | - | - | - | 14.27 |
| **St. 12** | - | - | - | - | - | - | - | - |
| **St. 13** | - | - | - | - | - | - | - | - |
| **St. 14** | - | - | - | - | - | - | - | - |
| **St. 15** | - | - | - | - | - | - | - | - |
| **St. 16** | - | - | - | - | - | - | - | - |
| **St. 17** | - | - | - | - | - | - | - | - |

**Table 5: Q-table for the manufacturing system example.**

The symbol "-" shows the transitions that are not possible to be triggered. This helps to identify the most valuable actions on each state, remembering that the agent can only choose between the values of the two first columns, which represents the controllable actions.

According to this Q-table, in state 0, the most valuable action to be chosen is $a_1$. In fact, this is the only possible action to be taken. Action $a_2$ would be also possible, but it has been disables by control (see restriction model $R$). In state 3, there are 2 possible actions: $b_1$

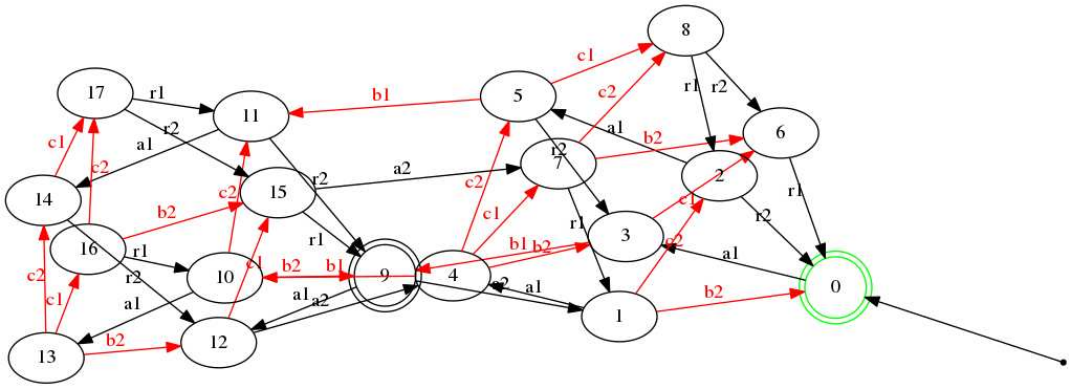

Figure 12: FSM modeling $K$, the input to the Gym environment.

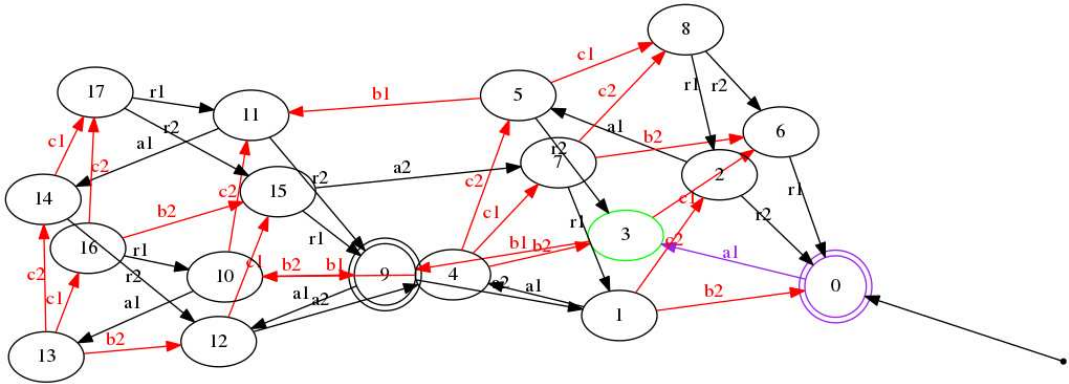

Figure 13: Automaton K's state $S_4$ represented in Gym environment

and $c_1$, but both are uncontrollable, and can not be taken by the control agent, even knowing that an action $b_1$ would return a better profit. In state 9 there are 2 controllable actions to be chosen: $a_1$ and $a_2$, in which action $a_2$ has a higher Q-value because it gives an immediate profit of 10, while choosing $a_1$ will consequently give this reward, but not immediately.

We further show the Q-table in Table 6, generated by the algorithm if we set to 100% the probability for machine 1 to crash (event $c_1$). That is, action $c_1$ occurs every time the FSM is in a state where it is eligible. In this case, the FSM always makes the path through the states 0, 3, 6, 0, which is in fact the only way possible, under this assumption. Note that the table only shows values different from zero in 3 cells, corresponding to transitions from state 0 to 3, 3 to 6 and 6 to 0. In summary, the table shows that, under this assumption, it does not matter which actions to choose, as the agent will always have a loss.

We can conclude that this example was modeled, converted to an MDP environment, and trained via RL algorithm successfully. Next, we further present second case study.

## 6.2 Transmitters Sharing a Channel

This example was previously shown in Section 3.4, in which two transmitters $T_1$ and $T_2$ sharing a communication channel $C$ that supports only one communication request at a time, as shown in Figure 14.

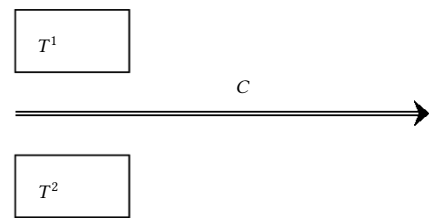

Figure 14: Example of a concurrent transmission system.

*6.2.1 Modeling of the system.* : the step-by-step modelling of this example was already shown in section 3.4, and therefore here we concentrate the focus on the final composition $K_T$ in Figure 16. Plant ans restrictions are repeated in Figure 15.

By composing $G_{T_1}$, $G_{T_2}$ and $R$, we obtained the FSM $K_T = G_{T_1} \| G_{T_2} \| R$, with 8 states and 14 transitions, shown in Figure 16.

*6.2.2 Conversion from DES to MDP:* To convert the system to an MDP, the FSM $K_T$ was exported to a XML structure and parsed through gym's methods to extract the DES information. Since the

|       | $a_1$  | $a_2$ | $b_1$ | $b_2$ | $c_1$  | $c_2$ | $r_1$  | $r_2$ |
|-------|--------|-------|-------|-------|--------|-------|--------|-------|
| St. 0 | -13.56 | -     | -     | -     | -      | -     | -      | -     |
| St. 1 | -      | -     | -     | -     | -      | -     | -      | -     |
| St. 2 | -      | -     | -     | -     | -      | -     | -      | -     |
| St. 3 | -      | -     | -     | -     | -14.69 | -     | -      | -     |
| St. 4 | -      | -     | -     | -     | -      | -     | -      | -     |
| St. 5 | -      | -     | -     | -     | -      | -     | -      | -     |
| St. 6 | -      | -     | -     | -     | -      | -     | -12.61 | -     |
| St. 7 | -      | -     | -     | -     | -      | -     | -      | -     |
| St. 8 | -      | -     | -     | -     | -      | -     | -      | -     |
| St. 9 | -      | -     | -     | -     | -      | -     | -      | -     |
| St. 10| -      | -     | -     | -     | -      | -     | -      | -     |
| St. 11| -      | -     | -     | -     | -      | -     | -      | -     |
| St. 12| -      | -     | -     | -     | -      | -     | -      | -     |
| St. 13| -      | -     | -     | -     | -      | -     | -      | -     |
| St. 14| -      | -     | -     | -     | -      | -     | -      | -     |
| St. 15| -      | -     | -     | -     | -      | -     | -      | -     |
| St. 16| -      | -     | -     | -     | -      | -     | -      | -     |
| St. 17| -      | -     | -     | -     | -      | -     | -      | -     |

**Table 6: Q-table for the manufacturing system example under the assumption that a break is 100% certain to occur.**

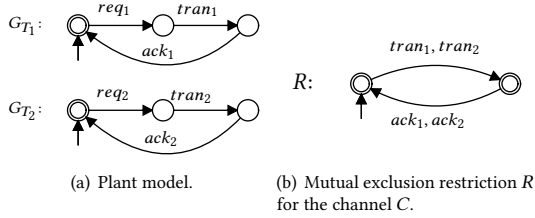

(a) Plant model.   (b) Mutual exclusion restriction $R$ for the channel $C$.

**Figure 15: Example of a concurrent transmission system.**

FSM has 8 states, it is possible to see it straightforwardly by calling function *render* of the gym environment. The function returns the image presented in Figure 16.

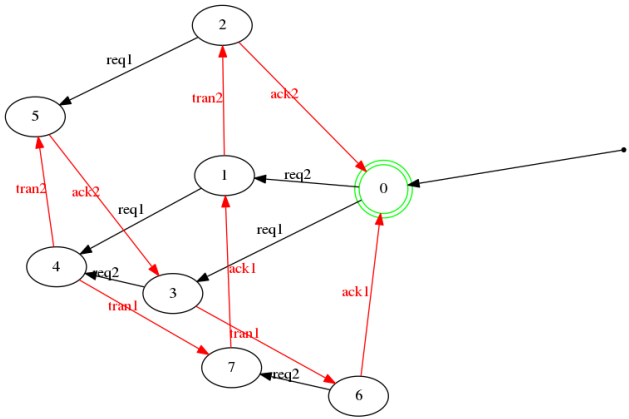

**Figure 16: Automaton $K_T$ for transmitters example.**

Kallil M. C. Zielinski, Marcelo Teixeira, Richardson Ribeiro, and Dalcimar Casanova

It worth remembering that the black transitions are triggered by controllable events and the red ones are triggered by uncontrollable events.

The list of rewards and triggering probabilities adopted for each action in the system is shown in Table 7. To differentiate the use of each transmitter, we intentionally consider that rewards for the *ack* signals are distinct. While $ack_1$ receives a reward of 2, $ack_2$ receives a reward of 3, indicating that, by having the choice prerogative, the RL algorithm should prefer to choose transmitter $T_2$.

| Action  | Reward | Probability set $P$ (%) |
|---------|--------|-------------------------|
| $req_1$ | -1*    | not applicable          |
| $req_2$ | -1*    | not applicable          |
| $tran_1$| -1*    | not specified           |
| $tran_2$| -1*    | not specified           |
| $ack_1$ | 2      | not specified           |
| $ack_2$ | 3      | not specified           |

**Table 7: Rewards adopted for the two transmitters example.**

*6.2.3 Applying the algorithm and review of the results.* For this case study, we use the Deep Q algorithm and also consider the rewards and transition probabilities from Table 7. The used neural network consist of an embedding layer as input, in which the state enters the neural network and is encoded in 10 distinct values. In sequence, there are 3 more fully connected layers with 50 neurons each. The output layer consists in a number of output neurons equal to the number of possible actions in the environment, 6 in this case.

The input and output of the used Deep Q network, is represented in Figure 17 and shows that the Deep Q network receives the current state $s_i$, which can be any number between 0 and 7 (see Fig. 16), as input, and processes this state to get the values for all Q-values related to the state $s_i$, which in this case are 6, related to the action space.

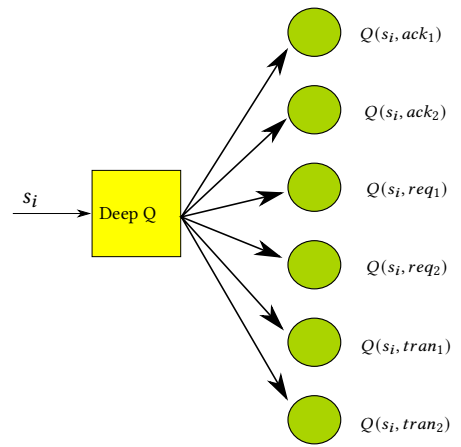

**Figure 17: Deep Q system for the transmitters example.**

Upon training for 100 episodes, the resulting Q-table is shown in Table 8.

|       | $ack_1$ | $ack_2$ | $req_1$ | $req_2$ | $tran_1$ | $tran_2$ |
|-------|---------|---------|---------|---------|----------|----------|
| **St. 0** | 0.01  | -0.02 | 5.28  | 6.67  | 0.01  | -0.04 |
| **St. 1** | 0.04  | -0.00 | 6.36  | 0.05  | -0.02 | 7.75  |
| **St. 2** | -0.04 | 8.85  | 6.94  | -0.01 | 0.04  | -0.04 |
| **St. 3** | -0.03 | -0.01 | -0.04 | 6.36  | 5.87  | 0.02  |
| **St. 4** | -0.01 | -0.05 | 0.02  | 0.00  | 7.46  | 7.16  |
| **St. 5** | -0.03 | 9.29  | 0.03  | 0.04  | 0.02  | -0.04 |
| **St. 6** | 6.64  | 0.00  | 0.02  | 6.94  | -0.00 | -0.02 |
| **St. 7** | 9.67  | -0.04 | -0.02 | -0.05 | 0.02  | -0.03 |

**Table 8: Q-table representing the training for the 2 transmitters example.**

Note that there is only a few values in the table that are not close to zero. These values correspond to the transitions that the FSM cannot trigger. As the neural network does not differ between possible and impossible transitions, it continues to update its weights in order to minimize the loss function in Equation 6, and approximates these Q-values to zero, since impossible transitions have a value zero in the Q-table.

Also remark that the only state in which the agent chooses between two or more controllable actions is the state 0. It has the option to choose between actions $req_1$ and $req_2$, and since using the transmitter 2 impacts on receiving a reward of +3, a few steps forward with the uncontrollable signal $ack_2$, the Q-value for $req_2$ is higher, indicating that the agent prefer it.

## 7 FUTURE OPTIMIZATIONS

In this paper, we are considering that the probability distribution is uniform among all events. In forward studies we pretend to work on adapting these probabilities to other kinds of distributions, like exponential distribution. For example, in the first day (episode) a machine have 5% of chance of crashing, while in the second day has 10%, and in subsequent days this probability is going to get higher and higher.

Another important modification is to turn possible not only the parsing through XML structures, but also in other Supremica output formats, like WMOD. This will turn easier for researchers to simply export any structure of automaton to gym and begin their training without worrying about the file's structure.

There is also a possibility of implementing a neural network with variable neurons in the output layer. In case of Section 6.2 example, the implementation of the Deep Q network considered that for all the states in the environment, we have all possible actions to be taken, but when working with FSMs, there is only some actions that are possible to be taken in a single state, so, Table 8 showed us that the neural network used wasted some processing and space that is not necessary in this case. Also, on complex systems with a huge state space, larger neural networks are needed, and also requires more complex structures, like recurrent LSTM cells.

## 8 FINAL CONSIDERATIONS

This article aimed to develop a tool that solves problems related to industry 4.0, in which are possible to be modeled as discrete event systems using reinforcement learning methods.

The work not only explored the tool itself but also pointed out differences between RL and DES environment and focused on solving them in order to create methods capable of transforming DESs into RL environments. The similarity between both types of environments allowed the creation of an easy-to-use tool capable of optimizing DES models with training via RL.

The use of RL in DESs evidences practical appeal, considering that the reward system used in RL algorithms can reflect many aspects in 4.0 industry modeled as DESs, like the cost of using factory machines. On the other side, there is a profit if some machines produce well. Then, RL emerges as an alternative that can anticipate whether or not certain actions are attractive for the factory. The price to be paid is a minor engineering effort: engineers have to simply provide the input model constructed as a FSM, and a reward list for each triggered event in the system. They have also the option to add probabilities for uncontrollable events.

Finally, we believe that the paper and the tool have potential to serve as a foundation for future studies involving RL and DESs, specially in industry 4.0-aware scenarios.

## REFERENCES
[1] Martín Abadi, Ashish Agarwal, Paul Barham, and others. 2015. TensorFlow: Large-Scale Machine Learning on Heterogeneous Systems. http://tensorflow.org/ Software available from tensorflow.org.
[2] T. Bangemann, M. Riedl, M. Thron, and C. Diedrich. 2016. Integration of Classical Components Into Industrial Cyber-Physical Systems. *Proc. IEEE* 104, 5 (May 2016), 947–959.
[3] A. V. Bernstein and E. V. Burnaev. 2018. Reinforcement learning in computer vision. In *Tenth International Conference on Machine Vision (ICMV 2017)*, Antanas Verikas, Petia Radeva, Dmitry Nikolaev, and Jianhong Zhou (Eds.), Vol. 10696. International Society for Optics and Photonics, SPIE, 458 – 464. https://doi.org/10.1117/12.2309945
[4] Greg Brockman, Vicki Cheung, Ludwig Pettersson, and others. 2016. OpenAI Gym. arXiv:arXiv:1606.01540
[5] Christos G Cassandras and Stephane Lafortune. 2009. *Introduction to discrete event systems*. Springer Science & Business Media.
[6] R. Drath and A. Horch. 2014. Industrie 4.0: Hit or Hype? [Industry Forum]. *IEEE Industrial Electronics Magazine* 8, 2 (June 2014), 56–58.
[7] R. Harrison, D. Vera, and B. Ahmad. 2016. Engineering Methods and Tools for Cyber-Physical Automation Systems. *Proc. IEEE* 104, 5 (May 2016), 973–985.
[8] Daniel Hein, Stefan Depeweg, Michel Tokic, and others. 2017. A Benchmark Environment Motivated by Industrial Control Problems. (09 2017), 1–8.
[9] L. P. Kaelbling, M. L. Littman, and A. W. Moore. 1996. Reinforcement learning: A survey. *Journal of Artificial Intelligence Research* 4 (1996), 237–285.
[10] Henning Kagermann, Wolfgang Wahlster, and Johannes Helbig. 2013. Recommendations for implementing the strategic initiative INDUSTRIE 4.0. *Final report of the Industrie 4.0 Working Group* (April 2013), 1–82.
[11] Henning Kagermann, Wolfgang Wahlster, and Johannes Helbig. 2013. *Recommendations for Implementing the Strategic Initiative INDUSTRIE 4.0 – Securing the Future of German Manufacturing Industry*. Final Report of the Industrie 4.0 Working Group. acatech – National Academy of Science and Engineering, München. http://forschungsunion.de/pdf/industrie_4_0_final_report.pdf
[12] Kallilmiguel. 2019. kallilmiguel/automata_gym. https://github.com/kallilmiguel/automata_gym
[13] Jens Kober, J. Bagnell, and Jan Peters. 2013. Reinforcement Learning in Robotics: A Survey. *The International Journal of Robotics Research* 32 (09 2013), 1238–1274. https://doi.org/10.1177/0278364913495721
[14] Maxim Lapan. 2018. *Deep reinforcement learning hands-on : apply modern RL methods, with deep Q-networks, value iteration, policy gradients, TRPO, AlphaGo Zero and more*. Packt Publishing, Birmingham, UK.
[15] Yuxi Li. 2017. Deep Reinforcement Learning: An Overview. *CoRR* abs/1701.07274 (2017). arXiv:1701.07274 http://arxiv.org/abs/1701.07274
[16] Y. Liu, Y. Peng, B. Wang, S. Yao, and Z. Liu. 2017. Review on cyber-physical systems. *IEEE/CAA Journal of Automatica Sinica* 4, 1 (Jan 2017), 27–40.

[17] Jelena Luketina, Nantas Nardelli, Gregory Farquhar, and others. 2019. A Survey of Reinforcement Learning Informed by Natural Language. *CoRR* abs/1906.03926 (2019). arXiv:1906.03926 http://arxiv.org/abs/1906.03926

[18] Volodymyr Mnih, Koray Kavukcuoglu, David Silver, and others. 2013. Playing Atari with Deep Reinforcement Learning. arXiv:arXiv:1312.5602

[19] Volodymyr Mnih, Koray Kavukcuoglu, David Silver, and others. 2015. Human-level control through deep reinforcement learning. *Nature* 518, 7540 (Feb. 2015), 529–533. https://doi.org/10.1038/nature14236

[20] László Monostori. 2014. Cyber-physical Production Systems: Roots, Expectations and R&D Challenges. *Procedia CIRP* 17 (2014), 9–13. https://doi.org/10.1016/j.procir.2014.03.115

[21] OpenAI. [n.d.]. A toolkit for developing and comparing reinforcement learning algorithms. https://gym.openai.com/

[22] Adam Paszke, Sam Gross, Francisco Massa, and others. 2019. PyTorch: An Imperative Style, High-Performance Deep Learning Library. In *Advances in Neural Information Processing Systems 32*, H. Wallach, H. Larochelle, A. Beygelzimer, F. d'Alché-Buc, E. Fox, and R. Garnett (Eds.). Curran Associates, Inc., 8024–8035.

[23] Yassine Qamsane, Mahmoud El Hamlaoui, Tajer Abdelouahed, and Alexandre Philippot. 2018. A Model-Based Transformation Method to Design PLC-Based Control of Discrete Automated Manufacturing Systems. In *International Conference on Automation, Control, Engineering and Computer Science*. Sousse, Tunisia, 4–11.

[24] P.J.G. Ramadge and W.M. Wonham. 1989. The control of discrete event systems. *Proc. IEEE* 77, 1 (1989), 81–98. https://doi.org/10.1109/5.21072

[25] Ferdie F. H. Reijnen, Martijn A. Goorden, Joanna M. van de Mortel-Fronczak, and Jacobus E. Rooda. 2020. Modeling for supervisor synthesis – a lock-bridge combination case study. *Discrete Event Dynamic Systems* 1 (2020), 279–292.

[26] Stuart J. Russell and Peter Norvig. 2009. *Artificial Intelligence: a modern approach* (3 ed.). Pearson.

[27] André Lucas Silva, Richardson Ribeiro, and Marcelo Teixeira. 2017. Modeling and control of flexible context-dependent manufacturing systems. *Information Sciences* 421 (2017), 1 – 14.

[28] David Silver, Aja Huang, Chris J. Maddison, and others. 2016. Mastering the game of Go with deep neural networks and tree search. *Nature* 529, 7587 (Jan. 2016), 484–489. https://doi.org/10.1038/nature16961

[29] Phil Simon. 2013. *Too Big to Ignore: The Business Case for Big Data* (1st ed.). Wiley Publishing.

[30] Richard S. Sutton. 1991. Dyna, an Integrated Architecture for Learning, Planning, and Reacting. *SIGART Bull.* 2, 4 (July 1991), 160–163. https://doi.org/10.1145/122344.122377

[31] Richard S. Sutton and Andrew G. Barto. 2018. *Reinforcement Learning: An Introduction* (second ed.). The MIT Press. http://incompleteideas.net/book/the-book-2nd.html

[32] Richard S Sutton, Andrew G Barto, et al. 1998. *Reinforcement learning: An introduction*. MIT press.

[33] Prasad Tadepalli and Dokyeong Ok. 1996. H-learning: A Reinforcement Learning Method to Optimize Undiscounted Average Reward. (03 1996).

[34] Gerald Tesauro. 2002. Programming backgammon using self-teaching neural nets. *Artificial Intelligence* 134, 1-2 (Jan. 2002), 181–199. https://doi.org/10.1016/s0004-3702(01)00110-2

[35] Christopher J. C. H. Watkins and Peter Dayan. 1992. Q-learning. In *Machine Learning*. 279–292.

[36] Christopher J. C. H. Watkins and Peter Dayan. 1992. Q-learning. *Machine Learning* 8, 3-4 (May 1992), 279–292. https://doi.org/10.1007/bf00992698

[37] Michael Wooldridge and Nicholas R. Jennings. 1995. Intelligent agents: theory and practice. *The Knowledge Engineering Review* 10, 2 (1995), 115–152. https://doi.org/10.1017/S0269888900008122

[38] Kallil M. C. Zielinski, Marcelo Teixeira, Richardson Ribeiro, and Dalcimar Casanova. 2020. Concept and the implementation of a tool to convert industry 4.0 environments modeled as FSM to an OpenAI Gym wrapper. https://github.com/kallilmiguel/automata_gym

[39] Knut Åkesson, Martin Fabian, Hugo Flordal, and Robi Malik. 2006. Supremica - An integrated environment for verification, synthesis and simulation of discrete event systems. *Proceedings - Eighth International Workshop on Discrete Event Systems, WODES 2006*, 384 – 385. https://doi.org/10.1109/WODES.2006.382401